# Prediction Accuracy & Reliability: Classification and Object Localization under Distribution Shift

Fabian Diet[1*], Moussa Kassem Sbeyti[1,2], Michelle Karg[1]

[1]Continental AG, Germany
[2]Technische Universität Berlin, Berlin, Germany
`fabi.nk97@yahoo.de`,
`moussa.kassem.sbeyti@campus.tu-berlin.de`,
`michelle.karg@continental-corporation.com`

**Abstract.** Natural distribution shift causes a deterioration in the perception performance of convolutional neural networks (CNNs). This comprehensive analysis for real-world traffic data addresses: 1) investigating the effect of natural distribution shift and weather augmentations on both detection quality and confidence estimation, 2) evaluating model performance for both classification and object localization, and 3) benchmarking two common uncertainty quantification methods - Ensembles and different variants of Monte-Carlo (MC) Dropout - under natural and close-to-natural distribution shift. For this purpose, a novel dataset has been curated from publicly available autonomous driving datasets. The in-distribution (ID) data is based on cutouts of a single object, for which both class and bounding box annotations are available. The six distribution-shift datasets cover adverse weather scenarios, simulated rain and fog, corner cases, and out-of-distribution data. A granular analysis of CNNs under distribution shift allows to quantize the impact of different types of shifts on both, task performance and confidence estimation: ConvNeXt-Tiny is more robust than EfficientNet-B0; heavy rain degrades classification stronger than localization, contrary to heavy fog; integrating MC-Dropout into selected layers only has the potential to enhance task performance and confidence estimation, whereby the identification of these layers depends on the type of distribution shift and the considered task.

**Keywords:** Distribution Shift, Uncertainty Quantification, Deep Ensemble, Monte-Carlo Dropout, Object Classification, Bounding Box Regression

## 1 Introduction

When deploying deep neural networks to real-world applications, such as perception for autonomous driving, challenging scenes usually cause a decline in detection performance. These scenes are often characterized by conditions such as adverse weather, low visibility, rare objects, or unfamiliar traffic environments, e.g., regional differences in traffic signs, vehicle types, emergency and attention signals, or road layouts. Therefore, studying the robustness of CNNs under *distribution shift* is substantial for deploying CNNs safely in safety-critical applications.

---

* This work was conducted while F. Diet was an intern at Continental AG.

Uncertainty quantification (UQ) has attracted attention in recent years to identify those CNN predictions that are made with low confidence [1–3]. This so-called *uncertainty* or - given a probabilistic interpretation - *confidence* can be caused by indistinct image information or by insufficient model generalization, where high model performance is achieved only for parts of the underlying true data distribution. The former is referred to as aleatoric uncertainty, the latter as epistemic uncertainty [1, 4]. Epistemic uncertainty can be estimated by various approaches including MC-Dropout, MC-DropConnect, Ensemble, and Flipout [2, 5]. Ensembles are commonly employed as a benchmark for quantifying epistemic uncertainty, simultaneously enhancing task accuracy and robustness. MC-Dropout is more memory-efficient and has a small influence on task accuracy [2, 5, 6]. This work explores four MC-Dropout variants differing in their computational effort and their coverage of variations in feature- or object-level representations.

Within the field of perception for autonomous driving, uncertainty quantification has been applied to 1) *classification tasks* such as semantic segmentation [1], 2) *regression tasks* such as depth [7, 8] and optical flow estimation [6], and 3) *combined tasks* such as object detection [2, 3, 9, 10]. These works focus on introducing novel UQ methods for a specific task. Yet, extensive benchmarks for uncertainty quantification, including also its robustness to distribution shift, are commonly performed only for classification data such as CIFAR-10, MNIST, and ImageNet [11, 12]. A single benchmark compares different UQ methods for object detection; thereby, jointly evaluating classification and localization uncertainty [2]. Motivated by a missing benchmark for object localization, we define the AD-Cifar-7 dataset so that both classification and localization accuracy can be evaluated independently. Furthermore, related works [2, 3, 6–10] select a single network, which varies across the studies, for their comparisons - due to long training times of application networks. Therefore, a comparison of *different network backbones* is missing. For this reason, three backbones differing in architecture and runtime (ResNet-50, EfficientNet-B0, ConvNeXt-Tiny) are selected for the robustness analysis using the AD-Cifar-7 dataset.

Closest to our work is the UQ benchmark under *distribution shift* for the classification datasets MNIST, CIFAR-10, and ImageNet [11]. It is based on ResNets and uses image augmentations such as blur, noise, and saturation to generate test data with different corruption intensity [11]. For the automotive use case, a comprehensive distribution shift dataset exists only for semantic segmentation, covering a wide range of distortions, including rain, fog, and out-of-distribution (OoD) samples for simulated traffic data [13]. Motivated by these first steps towards benchmarking UQ methods and towards studying robustness under natural distributional shifts, the AD-Cifar-7 dataset has been curated from several public autonomous driving datasets (BDD100K [14], NuScenes [15], KITTI [16], and CADC [17]). It covers the following types of natural distribution shift: different recording setup, adverse real-world weather conditions, simulated rain and fog, corner cases (CC), and OoD objects, see Fig. 1. The focus lies on traffic-related object classes including vehicle, pedestrian, traffic sign, traffic light, bike, bus, and truck. The ID dataset is based on cutouts from BDD100K [14]. The distributional shift datasets have been collated from publicly available data and by simulating rain and fog.

This work is one of the first large studies investigating the effects of distribution shift on both classification and regression, considering both accuracy and confidence estimation, and going beyond artificial image manipulations towards natural distortions. We select a *granular approach* and divide the CNN representations, the distributional shift datasets, the application of MC-Dropout, the network, the task, and the evaluation into information granules: 1) considering low-level-feature, high-level-feature, and object-level representations to explain performance differences, 2) using eight types of dataset shift covering natural distortions and weather augmentations, 3) applying MC-Dropout to selected network layers, 4) using three network backbones differing in compute time and network architecture, 5) studying classification and regression independently for the same input data, and 6) evaluating a carefully selected set of metrics for task performance and uncertainty quantification. This granular perspective enables a novel in-depth analysis and provides insights into the performance deterioration caused by different types of distortions and their effect on different neural network architectures.

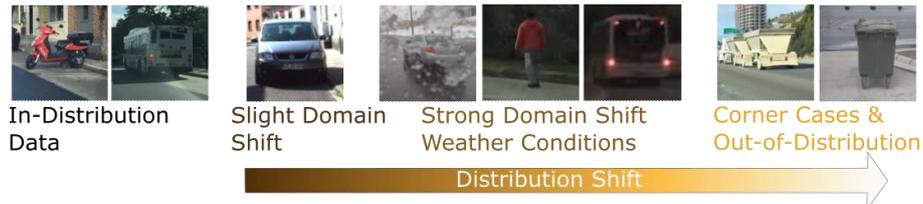

**Fig. 1.** Cropped images sourced from BDD100K [14], KITTI [16], CADC [17] and NuScenes [15] illustrating various types of distribution shift.

The contributions of this work are as follows: 1) a comprehensive analysis on the AD-Cifar-7 dataset shows that both the classification/regression accuracy and confidence estimation can *degrade significantly under natural and close-to-natural distribution shift*, e.g., under 80% accuracy and an *MAE* over 10*px* for heavy rain, 2) runtime-efficient MC-Dropout implementations depend on the *type of expected distortions*: feature-level distortions such as rain benefit from applying MC-Dropout layers to feature-level representations, object-level distortions such as corner cases and fog require applying MC-Dropout layers to the last layers of the network, 3) classification and regression rely to *varying degrees on representation granules* such that performance differences between UQ methods are observed: After-BB-Dropout is a very efficient UQ method for classification but not for regression across all types of distribution shift, 4) *ConvNeXt-Tiny is more robust* than ResNet-50 and EfficientNet-B0 across all UQ methods and distributional shifts, and 5) *carefully selected MC-Dropout layers are an efficient approximation* for Ensembles.

Section 2 provides background information and an overview of previous works for distribution shift and UQ. Section 3 summarizes the selected CNN backbones, CNN heads, and MC-Dropout variants. Section 4 provides an overview of the applied data curating process and describes the 9 datasets. The results are discussed in Section 5 and final conclusions are drawn in Section 6.

## 2 State of the Art

Given the importance of understanding and quantifying uncertainty for building robust and reliable models, a significant body of work exists on UQ methods in general. However, it is crucial to emphasize the importance of their analysis under distribution shift. A *distribution shift* occurs when the training and test data originate from different distributions, posing a challenge in generalizing from one to the other [18–20]. Such scenarios are inevitable when dealing with data gathered from the real world, which is inherently non-stationary. Consequently, they can manifest abruptly, evolve gradually, or follow seasonal patterns, triggered by specific events or due to a slowly changing environment [21].

Two significant factors for distribution shift are sample selection bias and non-stationary environments. Sample selection bias occurs when the training data is gathered unevenly from the underlying population, rendering it unreliable in representing the actual real-world operating conditions. On the other hand, non-stationary environments refer to scenarios, where the distribution undergoes changes over time. This may then lead to a deterioration in the model performance as the environment evolves [18, 22, 23]. These two factors can pose significant challenges, particularly in the dynamic environments of real-world applications such as autonomous driving, where distribution shifts are widely prevalent [11, 24].

Distribution shifts can be mainly classified into three types: covariate shift, label or prior probability shift, and concept shift. A graphical notation of the types of shifts and 12 additional sub-types of shifts can be accessed under the work of Kull and Flach [25]. *Covariate shift*, one of the most studied types of shifts [26], occurs when the input or source distribution changes between the training and the test data, while the conditional distribution between the input and output remains constant. An example of covariate shift can be observed when a model is initially trained using data collected in an urban city center but is subsequently tested on data originating from mountainous regions. *Prior probability shift* represents the opposite scenario, where the input features are assumed to be constant, but the distribution of the dependent variable (output or labels) is not. This type of shift arises when the prior assumptions made about the output distribution no longer hold, which is especially common in cases involving imbalanced datasets. Lastly, *concept shift* involves changes in both the input distribution and the relationship between the input features and the dependent target variables. In essence, it signifies that the extracted features of the model during training no longer align with those in the test data [18, 21]. In this work, we address all three types with varying weather conditions (mainly covariate shift), different datasets (mainly label shift), and corner cases (all three).

Despite a comprehensive understanding of the origins and variations in distribution shifts, specialized benchmarks are scarce for evaluating uncertainty quantification under distribution shifts [23]. Nevertheless, the evident challenges faced by current models in the presence of natural distribution shifts are well-studied. Moreno-Torres et al. [18], Taori et al. [22], Gustafsson et al. [27] investigate the robustness of classification models under natural distribution shift. They also mention, alongside Koh et al. [23], Gustafsson et al. [27], that most works primarily rely on synthetic distribution shift scenarios [11, 28–32]. These however may be informative, but also do not capture the complexities of real-world data distribution changes. Taori et al. [22] find most models provide no robustness and there is little to no correlation between natural and synthetic shifts. Furthermore, much of the literature on UQ, particularly under distribution shift, focuses on classification tasks. However, regression models suffer equally under distribution shifts, and many works target the increase of their robustness [12, 33–38]. Therefore, our work, combining both classification and regression, aims to bridge the gap between these two tasks. In doing so, we lay the groundwork for tackling more complex tasks such as object detection and unify the understanding of models and their uncertainty under distribution shift, regardless of the specific task at hand.

One of the few previous works on uncertainty analysis under distribution shift is presented in [11, 24]. They investigate how different methods such as Ensemble, MC-Dropout, and a variant of MC-Dropout, Last-Layer-Dropout (LL-Dropout), where dropout is only applied on the activations before the last layer, behave under synthetic distribution shift in a classification task. They find that deep Ensembles, with five Ensembles being sufficient, outperforms the other methods and is most robust to dataset shift. LL-Dropout, in comparison to full dropout, computes a more calibrated uncertainty under shift. Furthermore, post-hoc calibration works well assuming identically distribution but fails under even a mild shift in the input data. They find that the quality of the uncertainty degrades with the strength of the shift. A similar benchmark of uncertainty quantification methods under synthetic shifts is presented by [32] but for regression. Their evaluation of MC-Dropout and Ensemble alongside Bayesian models resulted in no conclusive ranking, especially not under a single metric. They find, however, that Ensemble and MC-Dropout return conservative uncertainty estimates under shift. This overconfidence is also discovered by [27] in their benchmark of regression UQ methods under synthetic and real-world distribution shift. Comparing identical UQ methods to [32], [12] propose another classification benchmark under synthetic distribution shift. They find Ensemble to slightly outperform the other methods, but none of the methods succeed in reporting a reasonable increase in uncertainty under distribution shift. Instead, the models increase their confidence with the severity of the shift, similar to [11]. Therefore, they reiterate the importance of measuring the quality of the prediction uncertainty alongside the model, particularly in safety-critical applications. In general, even on In-Distribution data, Ensembles typically exhibit superior performance compared to competing methods [39], albeit at the cost of a higher computational and memory complexity [11, 12]. Also close to our work are the benchmarks introduced by [23, 40, 41], which present a collection of datasets with naturally occurring distribution shifts. All these works are fundamental in facilitating the analysis of the safety of models and their predicted uncertainty when faced with distribution shifts. Therefore, we build on the latter to introduce a novel benchmark for the real-world application of autonomous driving for both classification and regression tasks, evaluating multiple networks with different uncertainty estimation methods under both weather augmented and natural distribution shifts.

## 3  Background and Methods

As mentioned in Sec. 2, simultaneously investigating regression and classification tasks within the same context is crucial for a more general understanding of uncertainty and model behavior under distribution shift. This is particularly vital given that a majority of real-world applications necessitate models that support both tasks. Furthermore, it is also imperative to investigate various models and UQ methods to attain a higher degree of generality. This section offers essential details regarding our selection of tasks, baseline models, and uncertainty estimation methods, including foundational knowledge of each.

### 3.1  Classification and Regression

Our baseline models are implemented with a regression and a classification head.

**Classification.** The classification head comprises of two fully connected layers containing $n$ neurons and a ReLU activation. It is followed by a third dense layer with the number of units equal to the classes, using linear activation for computing logits, and ending in a final softmax layer. During training, the weights are optimized using the cross-entropy loss.

**Regression.** Like classification, the regression head is composed of two fully connected layers, each with $n$ neurons and a ReLU activation. It also incorporates a third dense layer with four units to handle the four localization variables, utilizing linear activation. We use the Mean-Absolute-Error (MAE) loss for training.

The interpretation of localization variables varies, particularly within different object detection frameworks. Following the conventions of the most prevalent frameworks, we explore three different implementations of the localization head, as illustrated in Figure 2. The *Corner Points* approach describes bounding boxes using the coordinates of the top-left $(x_1, y_1)$ and bottom-right $(x_2, y_2)$ corners, applied for example in CornerNet [42]. The *Center Point* approach defines bounding boxes via the object center $(C_x, C_y)$, height $h$, and width $w$; as implemented in CenterNet [43]. The *Anchor-based* approach consists of relative size and position of bounding boxes to a predefined reference bounding box, i.e., anchor, used in detectors such as SSD [44] and EfficientDet [45]. For example, the average of all bounding boxes in the training data can be used as a pre-defined anchor. Similar to the Center Point, the network predicts four offsets for the center $(t_x, t_y)$, height $(t_h)$, and width $(t_w)$ in relation to the anchor. The final bounding box coordinates can then be calculated, for example, using [46, 47]:

$$w = e^{t_w} \cdot w_a, h = e^{t_h} \cdot h_a, c_x = t_x \cdot w_a + c_{x_a}, c_y = t_y \cdot h_a + c_{y_a}.$$

The architecture of the network and the number of neurons in the output layer do not differ between the three variants. Only the definition of labels for the training data is adjusted.

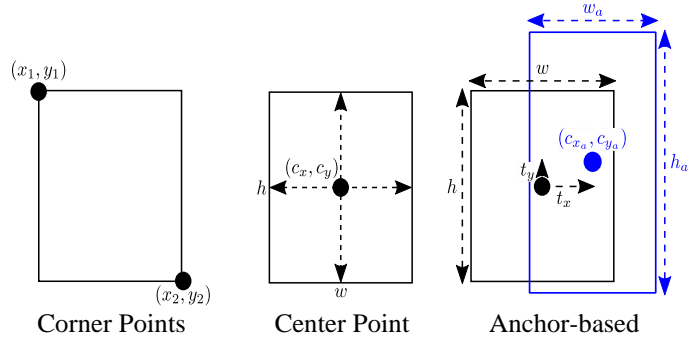

**Fig. 2.** Illustration of the three unique bounding box definitions for object localization: Corner Points, Center Points, and Anchor-based.

### 3.2 Baseline Models

In recent years, several novel network architectures have been introduced. This work examines three exemplary network architectures, chosen for their distinct characteristics in terms of parameter count, runtime, and performance. These are ResNet-50, EfficientNet-B0, and ConvNeXt-Tiny.

**ResNet.** *Residual Networks* (ResNet) [48] are characterized by a stack of residual blocks that incorporate skip connections and a bottleneck structure. The first is specifically designed to allow a transfer of information across the multiple layers of the network, hence addressing the vanishing gradient problem and enabling the training of deeper networks. The second improves computational efficiency and reduces the number of parameters in the network. The network structure is illustrated in Fig. 3. ResNet-50 is a 50-layer variant that significantly improves accuracy at 74,9% on ImageNet [49, 50] and is a foundation for various object detection models such as Mask R-CNN [51].

**EfficientNet.** *EfficientNet* [52] is a family of eight models (EfficientNet B0 - B7) with increasing complexity and performance. It is founded on Mobile Inverted Bottlenecks (MBConv) introduced in MobileNetV2 [53], as depicted in Fig. 4. In contrast to ResNet, EfficientNet uses a 3x3 depthwise convolution and a swish activation function, replacing the final activation with a linear one to mitigate potential information loss caused by ReLU. It also incorporates a squeeze-and-excitation (SE) layer and a fusion of an average pooling and a dense layer to enhance performance with a minimal increase in parameters. Due to its superior performance compared to ResNet at 77.1% on ImageNet [49] and significantly fewer parameters (5.3 vs. 25.6 million) [50], we select EfficientNet as a representative model within the category of computationally efficient networks suitable for real-time applications.

**ConvNeXt-Tiny.** *ConvNeXt* [54] represents a novel network architecture that blends the strengths of traditional CNNs and vision transformers. It incorporates design principles from the Swin transformers [55] into ResNet, including a distinct bottleneck structure featuring depthwise convolution and a larger 7x7 kernel compared to ResNet. Batch normalization and ReLU are replaced with transformer-style layer normalization and GELU activation, reducing the number of normalizations and activations. Furthermore, ConvNeXt uses separate downsampling layers with normalization, contributing to training stability. It outperforms the Swin transformer on ImageNet [49] in terms of accuracy while retaining the simplicity and efficiency of CNNs [54], achieving a top-1 accuracy of 81.3% for the smallest variant [50] out of the five variants, illustrated in Fig. 5. We, therefore, also select ConvNeXt as a representative model within the category of computationally efficient networks suitable for real-time applications. Additionally, it serves as a representative example of design principles derived from transformers.

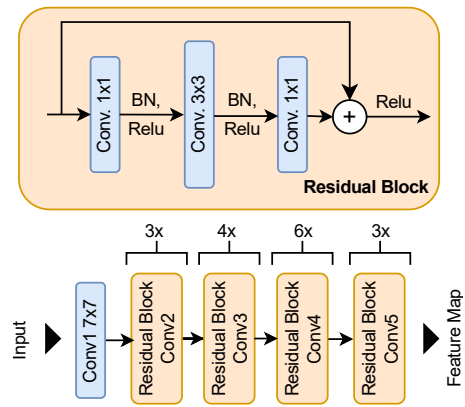

**Fig. 3.** Diagram of ResNet-50. Top: A residual block featuring a bottleneck structure along with a skip connection. Bottom: Network architecture consisting of a stack of residual blocks; pooling layers are not displayed.

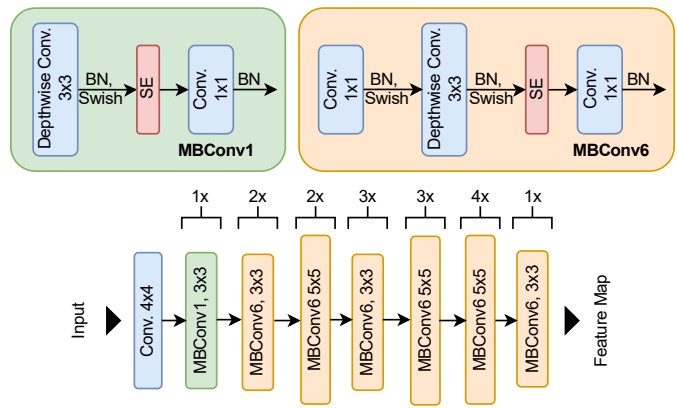

**Fig. 4**. Diagram of EfficientNet-B0. Top: Structure of MBConv blocks with 3x3 depthwise convolution, swish activation, and a squeeze-and-excitation layer. Bottom: Network architecture featuring a stack of MBConv Blocks.

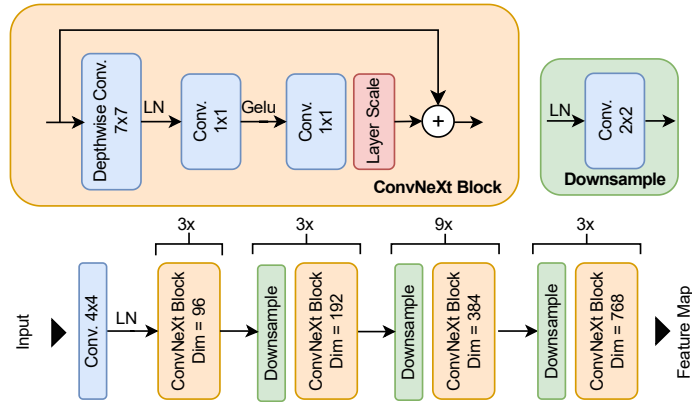

**Fig. 5.** Diagram of ConvNeXt-Tiny. Top: Structure of ConvNeXt and downsampling blocks featuring a bottleneck structure with a 7x7 depthwise convolution and a 2x2 convolution, respectively. Bottom: Network architecture consisting of a stack of both ConvNeXt and downsampling blocks. Adapted from [56].

**Hyperparameter Tuning.** *Hyperparameters* in neural networks are nontrainable parameters that influence the structure of the network, such as the number of hidden layers and activation functions, or its training process, such as learning rate and batch size [57]. Selecting their values carefully is essential for optimal network performance. However, manually finding the optimal hyperparameters is time-consuming and error-prone, leading to the introduction of various optimization strategies. *Grid Search* systematically explores all possible combinations of hyperparameters, while *Random Search* follows a similar strategy but randomly samples parameter values from predefined distributions [58]. Both methods are computationally intensive as they evaluate all parameter combinations. Another approach is *Hyperband* [59], which accelerates Random Search with adaptive resource allocation and early stopping, evaluating randomly selected parameter combinations with a minimal resource budget and continuing with the best candidates. On the other hand, *Bayesian Optimization* is a model-based approach aiming to find the global optimum with a minimal number of trials by predicting the likelihood of success for different parameter configurations [57]. We, therefore select Bayesian optimization due to its efficient resource utilization in the most promising configurations, saving computational time and resources. Furthermore, it is able to find the global optimum in the hyperparameter space, making it a robust choice for optimizing neural network configurations.

### 3.3 Monte-Carlo Dropout

*Dropout* [60] is a regularization technique for neural networks that mitigates overfitting by selectively deactivating individual neurons during training. For each neuron in the network layer where dropout is applied, a sample is drawn from the Bernoulli distribution to determine if it is dropped based on a pre-defined threshold. This threshold is a hyperparameter referred to as the *dropout rate*. The configuration of active neurons is then used to update the model weights during backpropagation.

Gal and Ghahramani [61] expand the application of dropout to estimate epistemic uncertainty using *Monte-Carlo Dropout* (MC-Dropout), a technique applicable to representations at various levels, including low, mid, high, and object levels. They demonstrate that the repeated prediction of input data with different dropout configurations active during prediction corresponds to the approximation of Bayesian inference. The epistemic uncertainty is therefore calculated as the variance over the $T$ repetitions of each prediction $p(y|x, \Theta_t)$ with a different dropout configuration on all input data [2, 61]:

$$p(y|x, D) \approx \frac{1}{T}\sum_{t=1}^{T} p(y|x, \Theta_t)$$

with the training data of $N$ samples $D = \{x_n, y_n\}_{n=1}^{N}$, the approximated predictive distribution $p(y|x, D)$ for a data sample $x$ from the input domain $X$, the ground truth $y \in \{1,...,C\}$ for classification with $C$ classes or $y \in \mathbb{R}^4$ for localization, and a sample of the network weights $\Theta_t$ with dropout from the weights domain $W$.

In practice, MC-Dropout serves as a straightforward technique for estimating the epistemic uncertainty. Adding dropout across the whole network affects low-level-feature, mid-level-feature, high-level-feature, and object-level representations. Nevertheless, its computational intensity due to the need for multiple stochastic passes can present challenges when applied in real-time scenarios. Additionally, as a hyperparameter, an improper choice of the dropout rate may result in poorly calibrated uncertainty estimates, prompting the need for an extensive hyperparameter search for optimal performance, as discussed in [62]. Therefore, as computationally efficient alternatives to full MC-Dropout (dropout both in the head and the backbone), this work also explores Last-Layer-Dropout (LL-Dropout), After-Backbone-Dropout (After-BB-Dropout), and Head-Dropout. The three alternatives are illustrated in Fig. 6. Each one affects representations at different stages of the network and serves various purposes. Therefore, their analysis offers a granular perspective on UQ.

In *LL-Dropout*, dropout is applied on the last layer before the output layer. This method primarily influences the object-level representations since it is implemented just before the final network output. Given the lower number of neurons in this layer compared to the entire network, LL-Dropout is computationally efficient.

*After-BB-Dropout* also utilizes a single dropout layer, but it is positioned directly after the network backbone, i.e., the initial layers responsible for feature extraction. This method influences the representations generated by the backbone before they are passed on to the head of the network. Therefore, After-BB-Dropout has a more significant impact on the high-level-feature representations compared to LL-Dropout.

*Head-Dropout* is a combination of the two previously mentioned variants. It involves applying dropout throughout the network head. By doing so, Head-Dropout provides a balance between influencing both high-level-feature and object-level representations in the network.

In the ResNet implementation of MC-Dropout, we adopt the approach detailed in [11], wherein a dropout layer is introduced following each ReLU activation within the Residual Block (as shown in Fig. 3). To maintain result comparability across the other two networks, we strive to minimize differences in their implementations. Specifically, for EfficientNet, we incorporate dropout after the Swish activation preceding the depthwise convolution (depicted in Fig. 4), and for ConvNeXt, it is introduced after the GELU activation (as shown in Fig. 5).

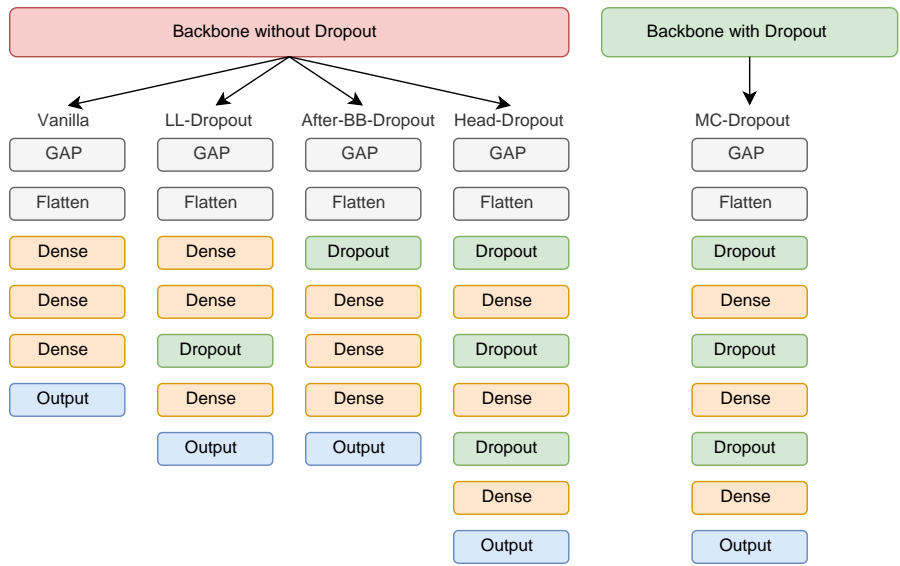

**Fig. 6.** Visualization of the different network layers, where MC-Dropout is applied.

### 3.4 Deep Ensemble

*Deep Ensemble* [63] represents an alternative approach for estimating epistemic uncertainty. Unlike MC-Dropout, which relies on a single model, Ensembles aggregate the predictions of multiple independently trained models sharing the same architecture. This approach not only enhances accuracy [64], but also results in a qualitative measure of uncertainty [11, 12, 39]. This can be attributed to their ability to identify distinct peaks within the parameter space, without being constrained to local uncertainties around a single peak [65]. In contrast to previous methods that often depended on bootstrapping, involving training models on different subsets of the training data, it has been shown that random weight initialization and the shuffling of training data alone are sufficient for accurate uncertainty estimation [63]. Therefore, similar to MC-Dropout, the epistemic uncertainty is extracted from the approximated predictive distribution of the $M$ networks in the ensemble and their corresponding weights $\{\Theta_m\}_{m=1}^{M}$ based on each prediction $p(y|x, \Theta_m)$ via: [2]

$$p(y|x, D) \approx \frac{1}{M} \sum_{m=1}^{M} p(y|x, \Theta_m)$$

Another advantage of this approach is its straightforward implementation, as it does not necessitate any modifications to the models themselves. However, a drawback lies in the substantial computational and memory resources needed for both training and prediction tasks across all models in the Ensemble. This becomes particularly challenging for deployments on edge devices in real-time, given that the computational complexity scales linearly with the number of models [2, 11].

## 4 Dataset Preparation

The datasets in this work are based on publicly available datasets. This chapter provides an overview of the data curation process and on relevant information for each of the datasets.

### 4.1 Motivation of the Dataset Design

In contrast to object detectors such as SSD [44] or EfficientDet [45], the separate classification and regression networks used in this work are limited to assigning only a single class and bounding box to each image. Common classification datasets, such as ImageNet [66] or MNIST [67], feature large and centered objects. However, real-world scenarios involve objects situated in various positions relative to the observing vehicle. To address this, we draw inspiration from the YouTubeBB dataset [68], which exhibits one or multiple objects of one class at a time in random positions in the images. We further constrain our dataset to only *one* main object per image, to align with the use of unified networks for both classification and regression. Given the specific focus on automotive applications, we concentrate on the core classes such as vehicles, pedestrians, traffic signs, and traffic lights. Additionally, the data reflects the diversity and authenticity of real-world conditions, encompassing a variety of different environments, weather conditions, and times of day.

### 4.2 Summary of the Methodology for Image Crop Selection

The dataset creation starts with images having at least one object and their associated bounding boxes. For each relevant object in the image, the center of the bounding box is calculated and virtually shifted in a random direction to ensure an even distribution of objects within the images. Subsequently, a 'cutout area' is defined based on the shifted center. While the size of the area is randomly chosen, it is guaranteed to be at least large enough to include the entire object. The objects are then cropped according to the cut-out area and uniformly scaled to a fixed size of 256x256 pixels, thereby eliminating the need for further rescaling during training. The coordinates of the bounding boxes are adjusted to align with the new image size and saved alongside the other annotations.

Since this process does not ensure that only one object is in the image at a time - as nearby objects may still fall within the cropped area - a pairwise comparison is conducted between all cutout areas and other objects in the original image. Images are classified as having a single main object only if the other objects occupy a maximum of 1/3 of the area of the main object. Various thresholds, such as 1/4 and 1/2, were tested, yielding inferior results. The scaling to 256x256 pixels introduces challenges, particularly for tiny objects in the original images, as they are significantly enlarged, leading to a loss in sharpness. To mitigate this, objects smaller than 30 pixels wide in the original images are removed, along with occluded or truncated objects.

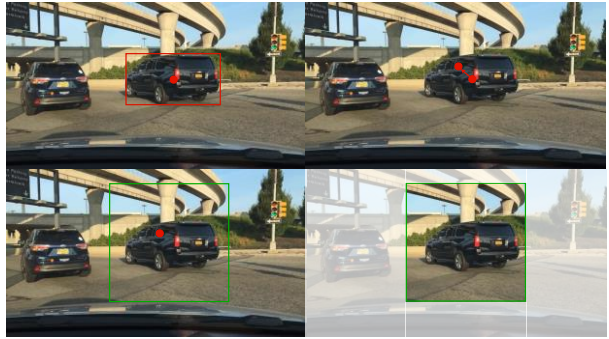

**Fig. 7.** Visualization of the cropping process on an image from BDD100K [14]. Upper left: original image with a bounding-box, upper right: virtually adjusted center point, lower left: definition of the cutout area, lower right: final cutout.

### 4.3 In-Distribution Dataset

The ID dataset serves as the basis for the following experiments. It consists of training, validation, and test data, see Tab. 1. We choose BDD100k [14] (BDD) as the starting point for our dataset curation. BDD is recorded in the regions of several US cities, such as New York, San Francisco, and Berkeley, and has a high diversity of street types, environments, weather conditions, and times of day, all captured in the annotations. In particular, BDD has previously served as a benchmark for uncertainty estimation [2], facilitating comparability with existing literature. To ensure scalability in our experiments, we set the dataset size according to CIFAR-10 [69]. The training set consists of up to 5000 images, while both the validation and test sets consist of up to 1000 cutouts per class to prevent significant class imbalance. The distribution of cutouts after filtering is shown in Tab.1.

The training and validation sets are extracted from BDD training set, and the test set is from the BDD validation set. The images with the BDD classes bicycle and motorcycle are combined into the new class bike due to the limited number of images. In addition, classes such as train are removed to maintain class balance, and rider is removed due to its inherent inclusion of both a person and a motorcycle. Truncated or occluded objects for the pedestrian and bicycle classes are not removed during the filtering process. Despite these measures, the specified 5000 or 1000 images cannot be collected for all classes, mainly caused by restricting each cutout to include only a single object. After filtering, all cutouts are manually inspected, and faulty objects are further removed. The high diversity of BDD is also evident in our datasets. Table 2 illustrates the distribution of cutouts across different weather conditions.

Table 1. Number of images for each class in the ID dataset.

| Class | Number of Images | | |
|---|---|---|---|
| | Train | Validation | Test |
| Traffic Sign | 5000 | 1000 | 1000 |
| Traffic Light | 5000 | 1000 | 1000 |
| Car | 5000 | 1000 | 1000 |
| Truck | 4281 | 1000 | 799 |
| Pedestrian | 3035 | 742 | 563 |
| Bus | 1732 | 431 | 309 |
| Bike | 1200 | 230 | 196 |

Table 2. Number of images for the different weather conditions.

| Class | Number of Images | | |
|---|---|---|---|
| | Train | Validation | Test |
| Undefined | 3295 | 756 | 641 |
| Clear | 12802 | 2654 | 2476 |
| Overcast | 3347 | 757 | 640 |
| Snow | 1950 | 442 | 411 |
| Partly Cloudy | 1922 | 392 | 363 |
| Rain | 1896 | 390 | 357 |
| Fog | 36 | 12 | 6 |

### 4.4 Distribution Shift Datasets

To explore the impact of distribution shift on uncertainty estimates, we generate several additional test datasets. As mentioned in Sec. 2, a simple form of distribution shift arises from the use of different sensors or different environments in which the data is recorded.

For our comparison, we utilize the KITTI Object Detection training dataset [16], which consists of 7841 images recorded in the region of Karlsruhe, Germany. One drawback of KITTI lies in the disparate definition of object classes compared to BDD, with only pedestrians and cars being identical between the two. Consequently, only these are considered for our tests. To create the dataset, we follow the same process as for the in-distribution data, resulting in a total of 1000 cutouts of cars and 517 of pedestrians.

We follow two approaches for real weather conditions: one dataset represents real images taken during snowfall and rain and one dataset represents simulated weather conditions. The real images are based on the Canadian Adverse Driving Conditions (CADC) dataset [17]. CADC is recorded in the Waterloo region of Canada, comprising 7,000 scenes from eight different cameras. All test drives are conducted in adverse weather, predominantly snowfall. To ensure comparability with the remaining datasets, we exclusively use images from the forward-facing camera. The CADC images are labeled as 3D-BB. For the conversion to 2D-BB of the camera coordinate system [70] is used. The conversion resulted in some BBs being significantly too large or incorrectly positioned. Therefore, 1000 images are manually re-labeled for each of the classes pedestrians and cars.

Simulated weather data is created by applying various filters of the Python library imgaug [71] to the in-distribution test dataset. Given the abundance of images of snowfall in CADC, our focus is directed toward simulating fog and rain conditions. Following the approach of Lakshminarayanan et al. [63], who evaluated distribution shift by rotating the images of the MNIST [11] dataset, we apply two levels of increasing rain and fog. It is worth noting that rain and fog filters produce varying effects on different images. The impact is less pronounced on brighter images compared to darker ones.

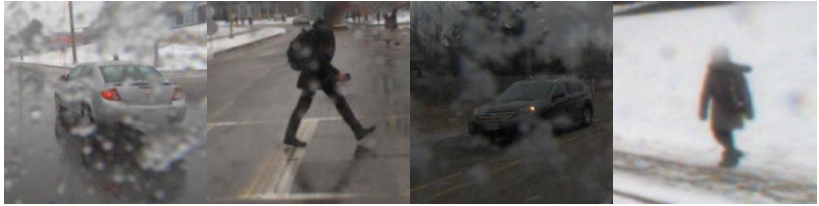

**Fig. 8**. Visualization of challenging weather conditions on cropped images from CADC [17].

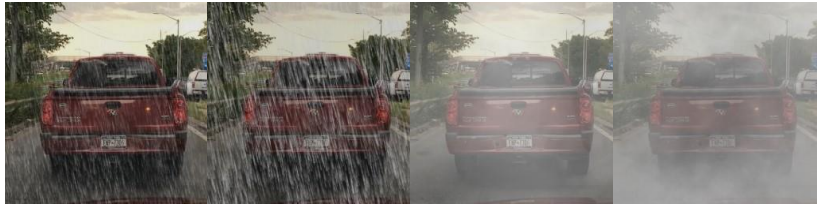

**Fig. 9**. Visualization of the four levels of simulated weather conditions applied on an image crop from BDD100K [14]. From left to right: light rain, heavy rain, light fog, and heavy fog.

### 4.5 Corner Cases and Out-of-Distribution Data

As mentioned in Sec. 2, distribution shift refers to changes in the distribution between training and testing data. Corner cases and OoD objects emerge when the model encounters instances during testing that differ significantly from those seen during training, contributing to performance challenges in these scenarios. A more precise definition varies depending on the specific application. In the scope of this study, data samples are categorized as corner cases when the parent class remains constant, yet the objects are not represented similarly in the training data. For instance, this includes construction vehicles such as concrete mixers, initially classified as trucks, or seated persons if the training data only includes standing persons. On the other hand, out-of-distribution samples refer to objects that bear no resemblance to those in the training data, such as traffic cones. In terms of classification, this implies that the object cannot be meaningfully assigned to any of the learned classes.

Two more datasets are created to investigate how UQ reacts to corner cases and out-of-distribution data. Both datasets contain images from BDD, KITTI and NuImages. NuImages is part of the NuScenes dataset [15]. It contains 93,000 images taken in the USA and Asia, and additional OoD classes, such as traffic cones and trash cans, which are not included in BDD or KITTI. Following our above definition, we define the two parent classes pedestrians and vehicles for the corner cases. Vehicles include all objects that can be assigned to one of the three classes car, truck or bus based on their general characteristics. This includes, for example, special vehicles such as concrete mixers, truck cranes, or trailers as an extended part of the vehicle class. Pedestrians are mainly people who differ from the training data due to special postures, such as sitting or bending. The out-of-distribution data includes objects, that cannot be assigned to any parent class of the training set but are real objects, that frequently occur in traffic. These are mainly stationary objects, such as traffic cones and garbage cans, but also trains and animals. A total of 438 images are selected for the corner cases and 500 images for the out-of-distribution data.

# 5 Experimental Results

We analyze the performance of the three baseline models (ResNet-50, EfficientNetB0, and ConvNeXt-Tiny) for the nine datasets. Hyperparameter tuning is performed on the validation set with KerasTuner [72]; final results are reported for the test set. The optimized hyperparameters are the number of neurons in the dense layers for the vanilla baseline, the learning rate, and the dropout rate. The number of repetitions is $M = 5$ for the Ensemble, recommended as sufficient in [2, 5, 63], and $T = 20$ for all MC-Dropout variants [2]. The CIFAR-10 policy from AutoAugment is used for image augmentation for classification and adapted for regression (by removing augmentations that modify the bounding box, because they reduce localization accuracy) [73]. First, evaluation metrics for task performance and UQ performance are summarized. Then, the performance on the ID data is discussed for classification and regression. It serves as a baseline in the following analysis, which covers the evaluation of the different networks under distribution shift, the robustness of UQ, and compute-efficient MC-Dropout variants.

## 5.1 Evaluation Metrics

Several metrics are used for evaluating UQ across different studies [2]. We select those metrics for task performance and uncertainty estimation, which are mostly used in related works [2, 11, 74, 74]. These criteria evaluate the confidence, sharpness, and correctness of a prediction.

Each image cutout $\mathbf{x_i} \in R^{256 \times 256}$ has a bounding box ground truth $y_{bb,i} \in \mathbb{R}^4$ and a class label $y_i \in \{1, \ldots, C\}$ with $C = 7$ for the in-distribution dataset and the four simulated datasets (Rain1, Rain2, Fog1, Fog2), and $C = 2$ (pedestrian, vehicle) for the KITTI, CADC, and CC dataset. The reduced number of classes for selected datasets affects only the results for classification, not for regression. The number of test samples $N$ is 4.867 for the ID and the simulated weather datasets, and 1.517, 2.000, 438, and 501 for the KITTI, CADC, CC, and OoD datasets, respectively.

The network weights $\Theta$ are trained for classification and regression independently. The classification head computes the *confidence score* $\hat{\boldsymbol{p}} \in \mathbb{R}^C$ and the predicted class $\hat{y} = \underset{c}{\operatorname{argmax}}\, p(\hat{y}_c | \boldsymbol{x}, \boldsymbol{\theta})$. The confidence scores range between [0,1]; the higher, the better. The head for the bounding box regression predicts $\hat{\boldsymbol{y}}_{bb} \in \mathbb{R}^4$ and the *uncertainty* $\hat{\boldsymbol{p}}_{bb} = p(\hat{\boldsymbol{y}}_{bb} | \boldsymbol{x}, \boldsymbol{\theta}) \in \mathbb{R}^4$. The uncertainty predictions range between $[0, inf[$; the lower, the better.

Our selection of the metrics for the classification is based on [11]:

— **Accuracy** (↑): The classification accuracy
$$\text{ACC} = \frac{1}{N} |\{\hat{y}_i \mid \hat{y}_i = y_i \text{ and } i = 1, \ldots, N\}|$$
evaluates only the correctness of a decision. The probability of a correct classification by chance is 14% for $C = 7$.

- **Expected Calibration Error** (ECE, ↓): The ECE is commonly used to evaluate the quality of UQ and calibration [11, 74]. It bins the predicted confidence values into $B$ bins each with $n_b$ samples, and computes for each bin $b$ the average confidence $\bar{p}_b = \frac{1}{n_b}\sum_{i=1}^{n_b}\hat{p}_{b,i}$ and the observed empirical accuracy

$$ACC_b = \frac{1}{n_b} | \{\hat{y}_{b,i} | \hat{y}_{b,i} = y_{b,i} \text{ and } i = 1, \dots, n_b\}|$$

  The difference between the predicted confidence $\bar{p}_b$ and the observed empirical accuracy $ACC_b$ is averaged over the bins:

$$ECE = \frac{1}{N}\sum_{b=1}^{B} n_b |ACC_b - \bar{p}_b|$$

  The ECE is an intuitive metric ranging between [0, 1]. The closer the predicted confidence to the observed empirical accuracy, the lower the ECE. It is visualized in the reliability diagram indicating over- and under-confidence for each bin $b$. In case of overconfidence, the predicted confidence estimates are higher than the observed empirical accuracy. In case of underconfidence, the predicted confidence is too low and the classification is more often correct than estimated by its confidence. The ECE has the advantage of an intuitive understanding, but it can be biased by insufficient or missing data points for some bins b and by the binning granularity. In our case, we choose $B = 10$ equidistant bins. Furthermore, the ECE does not evaluate the sharpness of a confidence vector, e.g., it does not distinguish between confidence vectors, where two or more classes are almost equally likely (e.g., p1 = [0.45, 0.44, 0.002, 0.002, 0.002, 0.002, 0.002]), and confidence vectors with a higher likelihood for one class and low likelihood for the other classes (e.g., p2 = [0.45, 0.07, 0.07, 0.07, 0.07, 0.07, 0.07]).
- **Negative Log-Likelihood** (NLL, ↓): The *NLL* is a proper scoring rule and evaluates the accuracy and sharpness jointly, and the predicted confidence of a prediction [75]. It ranges between $]-\inf, \inf[$. Its computation is based on individual samples and, hence, outliers can bias this metric.
- **Brier Score** (↓): The Brier score is also a proper scoring rule and evaluates accuracy, confidence, and sharpness jointly [75]. It ranges between [0,1] and computes the squared distance between predicted probability and 1 for the true class, using a one-hot encoded ground truth vector $y_n$ [76]:

$$BS = \frac{1}{N}\sum_{i=0}^{N}\sum_{c=1}^{C}(\hat{y}_{i,c} - y_{i,c})^2$$

Few works exist on metrics evaluating UQ for regression, some used only for specific use cases [6, 7, 47, 77]. Closest to our work are [47], who use the *NLL* for uncertainty quantification for bounding box regression, and [77], who formulate the *ECE* and a Platt-scaling-inspired calibration for regression. From these works, we select those metrics which are close to the ones for classification (see above):

- **Mean Average Error** (MAE [Pixel], ↓): The $MAE = \frac{1}{4N}\sum_i \sum_{j=1}^{4} |\hat{y}_{bb,ij} - y_{bb,ij}|$ evaluates the correctness of bounding box regression and is sensitive to outliers.
- **Intersection-Over-Union** (IoU, ↑): The IoU is commonly used in computer vision to assess the quality of the prediction of bounding boxes. It is the ratio of the intersection of the ground truth and the predicted bounding box divided by their union. It ranges between [0,1]. The IoU is 1, when bounding boxes are identical.
- **NLL** (↓): When assuming a Gaussian output distribution, the NLL is:
$$NLL = \frac{1}{4N}\sum_i \sum_{j=1}^{4} \frac{1}{2}\log(2\pi\hat{\sigma}_{ij}^2) + \frac{\hat{y}_{bb,ij} - y_{bb,ij}}{2\pi\hat{\sigma}_{ij}^2}$$
where the variance $\hat{\sigma}_{ij}$ is approximated by the predicted uncertainty $\hat{p}_{bb,ij}$. We consider only univariate models and do not model covariances. The NLL for regression evaluates the uncertainty and the correctness, yet not the sharpness. The sharpness for regression can be measured by the variance of $\hat{p}_{bb,ij}$. Similar to the NLL for classification, the NLL for regression can be affected by outliers.

- **ECE** (↓): The computation of the ECE is based on predicted confidences and observed empirical accuracies [78]. To apply the ECE computation to regression, [77] defines a set of confidence levels $p_l$ ranging between [0,1], in our case 10 equidistant levels l (same as for classification). For each confidence level $p_l$, the expected empirical frequency $\hat{p}_{emp,l}$ is computed:
$$\hat{p}_{emp,lj} = \frac{1}{N}|\hat{y}_{ij}|F_{ij}(\hat{y}_{ij}) \leq p_l, i = 1, ..., N| \text{ with } F_{ij}(\hat{y}_{ij}) = \frac{(\hat{y}_{bb,ij} - y_{bb,ij})}{\hat{\sigma}_{ij}}$$

It is computed for each bounding box coordinate $j$. The variance $\hat{\sigma}_{ij}$ is approximated by the predicted uncertainty $\hat{p}_{bb,ij}$. The interpretation of the ECE for regression is analog to classification, especially for the reliability diagram. The main difference is that for regression the ECE is based on confidence intervals for the predicted bounding box values [77].

### 5.2 Baseline Networks for ID (AD-Cifar-7)

**Classification Baseline.** The hyperparameter search revealed that the best number of neurons *n* is 128 for the two fully connected layers in the classification and the regression head. Tab. 3 shows the ID performance for classification for the vanilla networks. Commonly, an accuracy above 93% is achieved on the ID data. As some distribution shift subsets (KITTI, CADC, and CC) cover only 2 classes, we added the average of the classification metrics for pedestrians and vehicles for the ID data to Tab. 3. When only these two classes are considered, the baseline is slightly better.

Table 3. Task performance of the vanilla models for the ID dataset

|  | Classification | | | | Regression | |
|---|---|---|---|---|---|---|
|  | Accuracy ↑ | | NLL ↓ | | IOU ↑ | |
| No. Test Classes | 7 | 2 | 7 | 2 | - | - |
| EfficientNet-B0 | .947 | **.966** | **.156** | **.108** | 5.97 | .845 |
| ResNet-50 | .936 | .955 | .184 | .139 | 6.19 | .841 |
| ConvNeXt-Tiny | **.948** | .951 | .162 | .164 | **5.51** | **.858** |

**Regression Baseline.** The three approaches for defining the regression head, see Sec. 3.1, are compared using the ID data. Differences in performance between the Corner Points, Center Point, and Anchor-based approaches are relatively small. Similarly, the difference in using the MAE, MSE, IoU, or Distance-IoU as a loss function is negligible. The Corner Points approach and the MAE loss lead to consistently accurate bounding box estimation and, therefore, are selected as the baseline for the regression head. The results for regression are summarized in Tab. 3. The MAE is reported in pixels as units. When repeating the training for localization five times, the variance for the MAE is between .116−.145 and for the IOU 0.003 for the three networks. The variance between training runs is significantly smaller than the differences caused by a different network architecture.

### 5.3 Uncertainty Quantification for ID (AD-Cifar-7)

The learning rate and the dropout rate are optimized by a hyperparameter search for each network and each UQ method for classification and regression. The final learning rate $\lambda$ is 0.0001 for all classification networks, including their dropout extensions. For regression, this is only the case for ResNet-50 including its dropout variants and the vanilla EfficientNet-B0. The learning rate is a magnitude higher, 0.001, for the dropout variants of EfficientNet-B0 and all variants of ConvNeXt-Tiny. The dropout rate is consistent 0.05 for all regression networks. For classification, the dropout rate can be higher and varies between [0.05, 0.1, 0.15, 0.2, 0.25, 0.3, 0.35] with highest values for After-BB-Dropout and lowest values for MC-Dropout, see Tab. 4.

Table 4. Dropout rate for classification after a hyperparameter search.

|  | **EfficientNet-B0** | **ResNet-50** | **ConvNeXt-Tiny** |
|---|---|---|---|
| LL-Dropout | .1 | .2 | .05 |
| After-BB-Dropout | .3 | .35 | .15 |
| Head-Dropout | .05 | .1 | .05 |
| MC-Dropout | .05 | .05 | .05 |

**UQ for Classification.** The vanilla networks are extended by four variants of MC-Dropout and the results on the ID dataset are shown in Fig. 10. An Ensemble of 5 networks improves the accuracy significantly for all three networks and serves as a benchmark for the robustness evaluations. Depending on the chosen network, the accuracy differs slightly for the different dropout variants, with LL-Dropout often showing a performance below the vanilla baseline. Even though the accuracy differs only slightly for the vanilla ConvNeXt-Tiny, its extension by UQ achieves consistently better results than for ResNet-50 and EfficientNet-B0. The vanilla UQ for classification can be marginally improved by an Ensemble. MC-Dropout improves the ECE and the Brier score for ConvNeXt-Tiny and EfficientNet-B0, and After-BB-Dropout only for ConvNeXt-Tiny. This indicates a correlation between the accuracy of the baseline network and the performance of UQ methods.

**UQ for Regression.** For regression, all UQ methods achieve on average a high IoU. Similar to classification, better performance of the dropout variants is observed for the networks with higher localization accuracy, e.g., the localization accuracy of EfficientNet-B0 benefits slightly from dropout, whereas it often decreases for ResNet-50. As the vanilla networks for regression do not compute an estimate for uncertainty, we can compare the results only to our baseline, the Ensemble. Fig. 11 shows that LL-Dropout, Head-Dropout, and MC-Dropout are good candidates for computing uncertainties for localization. After-BB-Dropout leads to the largest decrease in accuracy for ResNet-50 and increases the ECE for the three networks. Hence, for ID data, MC-Dropout and its more compute-efficient variant Head-Dropout are good UQ methods for regression.

**Regression vs. Classification.** Studying classification and regression independently for the same data shows that they rely on different representations of the input data. Classification benefits more from After-BB-Dropout than bounding box regression indicating that high-level-feature representations are more relevant for classification. Regression seems to rely more on object-level representations and requires integrating dropout into the last dense layer of the regression head.

Note, that results have been shown for selected metrics with a focus on ECE, accuracy, and IoU. For NLL, we observe large values for After-BB-Dropout in all cases for regression, which are explained by outliers in the data.

**Towards Distribution Shift Analysis.** The results on the ID data lead to the following research questions for uncertainty quantification under distribution shift:

- How strong is the decrease in baseline performance under different types of distribution shift?
- How robust are the network architectures under distribution shift?
- Which compute-efficient dropout variants are more robust under distribution shift?
- Does uncertainty quantification for classification and regression rely on the same granule level of representations?
- Does distribution shift affect classification and bounding box regression in the same way?

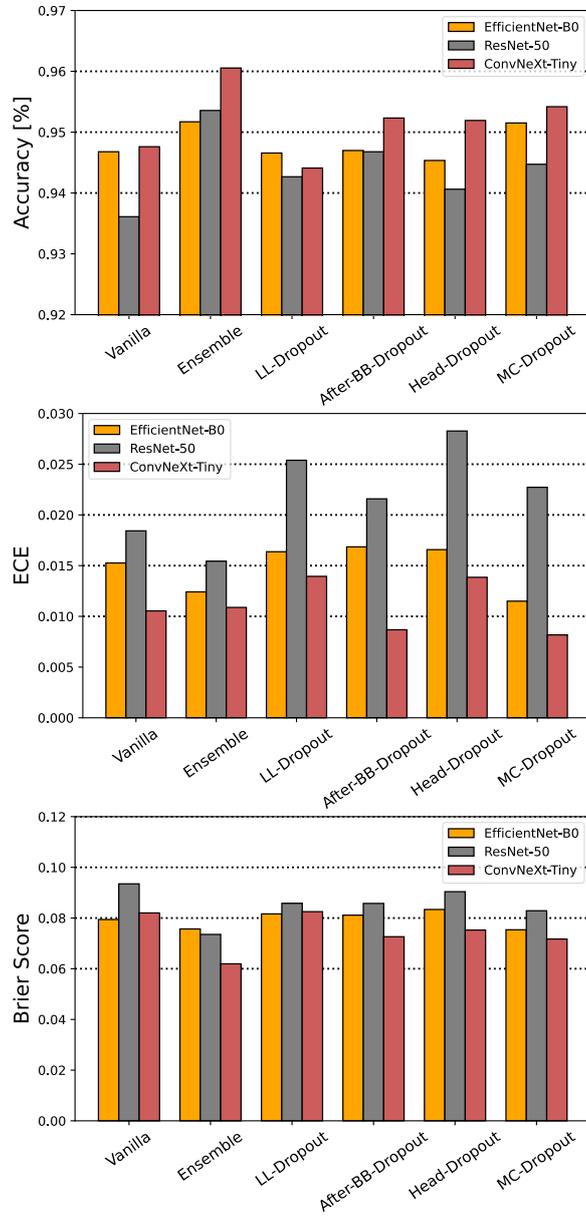

**Fig. 10.** Classification and uncertainty quantification are compared for the vanilla baselines, for Ensembles serving as the gold standard for our dataset, and for four different dropout implementations.

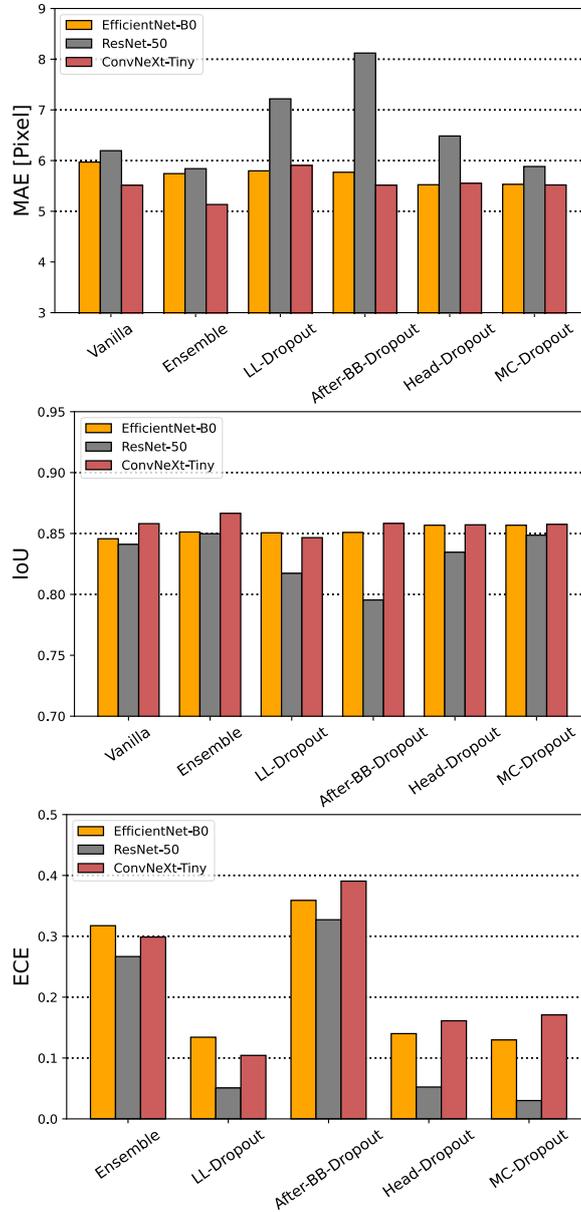

**Fig. 11.** Comparison between localization accuracy and uncertainty quantification for Ensembles serving as the gold standard for our dataset and for four MC-Dropout variants. (Note: the vanilla baselines for regression do not predict an uncertainty.)

### 5.4 Different Network Architectures under Distribution Shift

We investigate how different types of natural distribution shift influence the accuracy and ECE for different network architectures. For each metric, we accumulate the results of EfficientNet-B0, ResNet-50, and ConvNeXt-Tiny over the baseline and the UQ methods.

**Robust Classification.** Fig. 12 shows the performance decrease under the different types of natural distribution shift. Detection rates for KITTI are commonly higher than for the BDD dataset. Similarly, we observe a slight increase in classification performance for the KITTI subset, which cannot only be explained by the reduced number of classes. The classification is also reduced to two classes for the Real Weather and the Corner Cases dataset. Even though the classification rate by chance is higher in these cases, the performance drops in comparison to the ID baseline. Noticeable, is the *large drop in performance under heavy rain*, where the accuracy falls below 75% and the highest ECE values are observed. The performance drop for the Real Weather subset is less because it includes snow-covered winter scenes to a large extent. The Corner Cases dataset also shows the highest values for the ECE, but only an average decrease in accuracy. This indicates that, for corner cases, the confidence estimation is less robust than its class prediction. Overall, the difficulty of the natural distribution shifts can be sorted as follows: 1) changes in the camera setup lead to small performance changes, 2) natural occlusion of objects caused by snow or fog reduces the accuracy in the single-digit range, 3) corner cases can reduce accuracy by over 10%, and 4) rain and heavy fog lead to significant performance drops with accuracies below 85%. These observations are general across all networks and UQ methods.

The ranges in the boxplot Fig. 12 show that the ConvNeXt-Tiny leads to overall lower ECE values than EfficientNet-B0 and ResNet-50. This indicates that *ConvNext-Tiny is more robust to the investigated distribution shifts*.

**Robust Regression.** For regression, a decrease in performance is observed for all types of natural distribution shift; see Fig. 13. The evaluation for regression includes additionally the OoD set, for which bounding boxes are available; only class labels are missing. A stronger decrease in performance is observed for OoD samples in comparison to CC samples. Furthermore, the performance drop is similar for simulated light rain/fog and the Real Weather set. The largest decrease in performance is observed for heavy fog and heavy rain.

In comparison to classification, regression is more affected by heavy fog than heavy rain. Hence, *unclear object boundaries affect the bounding box localization stronger than the object classification.* Similarly to the results for classification, ConvNeXt-Tiny is more robust under distribution shift than ResNet-50 and EfficientNet-B0. In conclusion, we observe that 1) ResNet-50 is less robust to distribution shift than the more compute-efficient architectures EfficientNet-B0 and ConvNeXt-Tiny, 2) ConvNeXt-Tiny shows low performance variation when extended by different UQ variants, 3) the simulated Heavy Rain/Fog dataset leads to the strongest drop in performance.

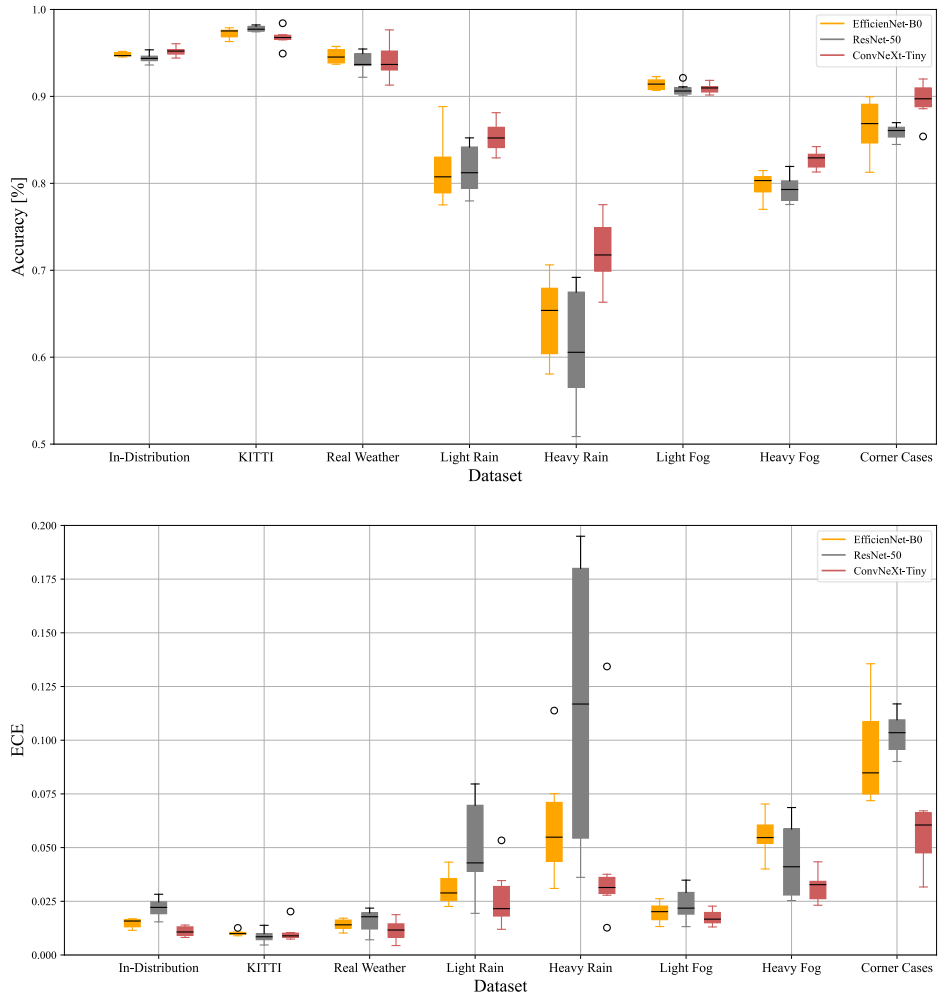

**Fig. 12.** Classification performance under distribution shift: simulated rain, heavy fog, and corner cases lead to a strong decrease in performance across all networks and UQ methods.

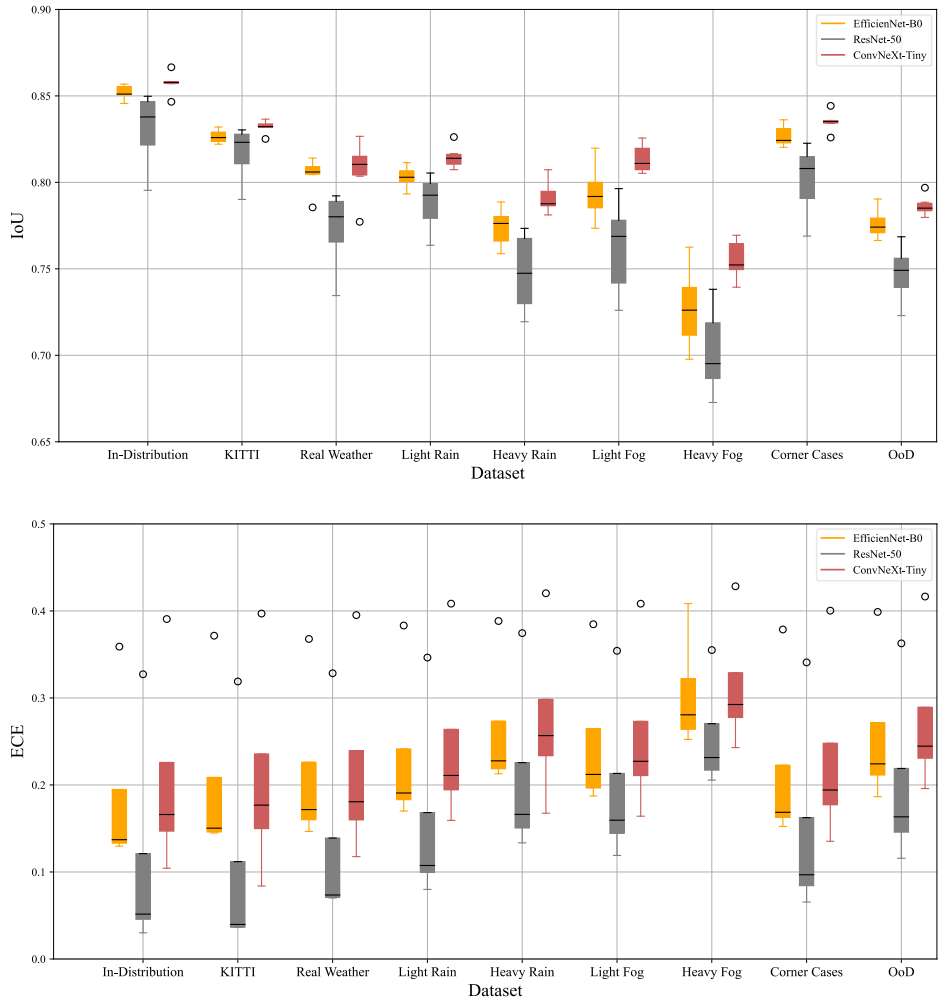

**Fig. 13.** Regression performance under distribution shift: a decrease in performance is observed for all types of distribution shift, strongest for heavy fog, heavy rain, and OoD.

## 5.5 Robustness of UQ under Distribution Shift

Classification and localization accuracy can significantly drop under natural distribution shift. Here, we analyze the robustness of MC-Dropout in comparison to Ensemble - widely considered as a gold standard for confidence estimation - under distribution shift.

**Robust UQ for Classification.** A noticeable improvement in classification accuracy is observed for Ensemble for all investigated types of natural distribution shift. MC-Dropout often achieves a slight improvement in classification accuracy, except for heavy rain and heavy fog. Similarly, confidence estimation is more robust to natural distribution shift for Ensemble, which is evident in both metrics ECE and the Brier Score, see Tab. 5. The ECE has the advantage that it can be computed for classification and regression and is less affected by individual outliers. The Brier score for classification and the NLL for regression are chosen as proper scoring rules, supporting our conclusions drawn from the ECE. Results are reported for ConvNeXt-Tiny because this network is consistently more robust under different types of distribution shift, see previous section.

**Table 5.** Classification accuracy and confidence estimation under distribution shift.

|  | ID | KITTI | CC | Weather | L. Fog | L. Rain | H. Fog | H. Rain |
|---|---|---|---|---|---|---|---|---|
| Acc. ↑ |  |  |  |  |  |  |  |  |
| ConvN. | .948 | .965 | .895 | .928 | .912 | .850 | .830 | .729 |
| +Ense. | **.961** | **.971** | **.913** | **.957** | **.918** | **.881** | **.842** | **.775** |
| +MC-D. | .954 | .968 | .854 | .936 | .909 | .854 | .816 | .697 |
| ECE ↓ |  |  |  |  |  |  |  |  |
| ConvN. | .011 | .020 | .057 | .010 | .017 | .019 | .033 | .032 |
| +Ense. | .011 | **.008** | **.044** | **.008** | **.015** | .018 | **.024** | **.013** |
| +MC-D. | **.008** | .009 | .067 | .015 | .016 | **.012** | .035 | .028 |
| Brier S. ↓ |  |  |  |  |  |  |  |  |
| ConvN. | .082 | .054 | .175 | .110 | .135 | .206 | .246 | .371 |
| +Ense. | **.062** | **.040** | **.136** | **.065** | **.119** | **.169** | **.217** | **.306** |
| +MC-D. | .072 | .047 | .238 | .094 | .133 | .200 | .256 | .388 |

**Robust UQ for Regression.** The vanilla CNN does not compute an uncertainty for localization. For bounding box regression, we observe that Ensembles provide a higher localization accuracy across the different types of distribution shift, whereas MC-Dropout computes more accurate uncertainty estimates for most types of distribution shift. The same is observed when using the NLL instead of the ECE as a confidence metric. An exception is the strong corruption of the input data by heavy rain; here, Ensembles provide a better uncertainty estimate, see Tab. 6.

In conclusion, these results show that MC-Dropout can improve robustness for classification and localization, but results vary depending on the type of distribution shift.

Table 6. Localization accuracy and uncertainty quantification under distribution shift.

|  | ID | KITTI | CC | Weather | L. Fog | L. Rain | H. Fog | H. Rain | OoD |
|---|---|---|---|---|---|---|---|---|---|
| IoU ↑ | | | | | | | | | |
| ConvN. | .858 | .832 | .817 | .815 | .836 | .823 | .784 | **.769** | .797 |
| +Ense. | **.867** | **.837** | **.826** | **.827** | **.844** | **.826** | **.797** | .768 | **.807** |
| +MC-D. | .857 | .832 | .815 | .804 | .834 | .805 | .783 | .739 | .788 |
| ECE ↓ | | | | | | | | | |
| +Ense. | .299 | .344 | .271 | .331 | .286 | .280 | .269 | **.281** | .267 |
| +MC-D. | **.171** | **.182** | **.206** | **.187** | **.197** | **.228** | **.242** | .296 | **.256** |
| NLL ↓ | | | | | | | | | |
| +Ense. | 12.6 | 33.9 | 14.6 | 13.3 | 11.4 | 10.9 | 10.6 | **11.1** | 9.3 |
| +MC-D. | **5.9** | **6.4** | **9.0** | **7.7** | **7.2** | **10.7** | **10.5** | 22.4 | **9.2** |

### 5.6 Robustness of Runtime-Efficient MC-Dropout Variants

Similar to the previous section, we choose ConvNeXt-Tiny for this analysis, which compares the performance of compute-efficient dropout variants with MC-Dropout.
In addition, only low dropout rates are selected by hyperparameter search for both regression and classification for ConvNeXt-Tiny, which simplifies the comparability of the results between classification and regression.

**Efficient UQ for Classification.** Fig. 14 shows the accuracy and the ECE for classification. Improved robustness under distribution shift can be achieved for After-BB-Dropout, where dropout is applied only to the fully connected layer after the feature extraction and covers variations in high-level-feature representations. Yet, applying dropout to more layers in the classification head, as for Head-Dropout, can result in a significant decrease in robustness, especially for strong distribution shift like light and heavy rain. The picture is slightly different for corner cases, where the distribution shift mainly affects object-level representations. Here, LL-Dropout can be beneficially. This is one of the first studies indicating that *CNN robustness can benefit from applying dropout to those network layers which are affected by the type of distribution shift.*

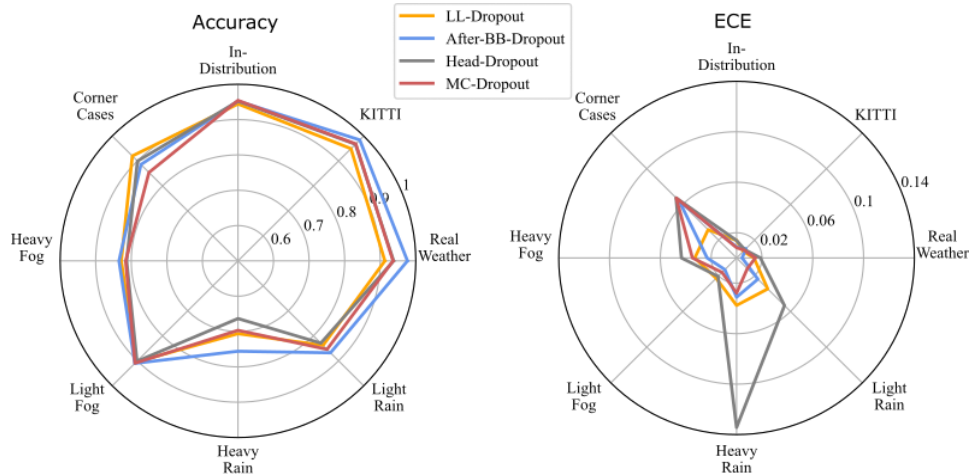

**Fig. 14.** For classification, a compute-efficient variant of MC-Dropout is applying dropout only to the fully connected layer after feature extraction, covering high-level feature representations (here shown for ConvNeXt-Tiny).

**Efficient UQ for Regression.** In the case of bounding box regression, see Fig. 15, the computational effort for MC-Dropout can be reduced to Head-Dropout without a loss of performance. In contrast to classification, After-BB-Dropout leads to a significant decrease in UQ for all types of distribution shift. This shows that uncertainty quantification for object localization requires dropout layers applied to the object representations in the CNN; applying dropout only to the feature representation is not sufficient for bounding box localization. Furthermore, LL-Dropout consistently improves the ECE for all types of distribution shift, but predicts in most cases less accurate bounding boxes than MC-Dropout.

**Types of Distribution Shift.** In summary, the performance of compute-efficient variants of MC-Dropout depends on the task (regression or classification), the selected dropout layers, and the type of distribution shift. For classification, MC-Dropout layers applied to high-level-feature representation is sufficient, whereas bounding box localization requires MC-Dropout layers applied to object-level representations. Distribution shifts, which are characterized by distortions to feature representations, such as light and heavy rain, can have a different effect on the robustness than distribution shifts, which distort mainly object-level representations, such as corner cases or OoD. *The efficiency of compute-efficient dropout variants capturing the variability in feature-level or object-level representations depends on the type of distribution shift.*

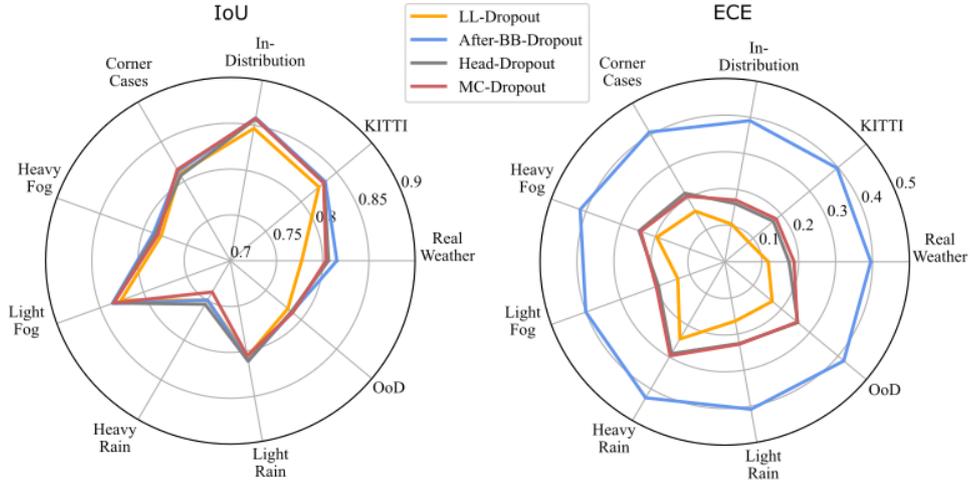

**Fig. 15**. For localization, uncertainty quantification is compute-efficient, when applying MC-Dropout to the last layer of the regression head, covering only object-level representations (here shown for ConvNeXt-Tiny).

## 6 Conclusion

This work takes a granular perspective on the robustness of CNNs and uncertainty quantification methods under distribution shift. In granular computing, complex problems are subdivided into smaller units, the granules, which can relate to each other hierarchically or horizontally [79, 80]. In our case, we subdivide a) the distribution shift data into granules, each representing a critical type of shift for real-world cases, b) the application of MC-Dropout to selected network layers only, and c) the learned representations of the CNN into low-level-feature, high-level-feature, and object-level representations serving as a linkage between the types of distribution shift and UQ methods. Granular computing is inspired by human problem solving and, by reducing complexity, it helps understanding complex mechanisms and designing efficient principles for machine computing [79–82]. Here, we make use of it for breaking down the complexity of the performance deterioration of CNNs under distribution shift into data granules, uncertainty granules, and representation granules. This approach allows us to gain a better understanding of the underlying principles causing this deterioration and improve the efficiency of MC-Dropout for CNNs. The key contribution is to relate the type of distribution shift and the selected method for uncertainty quantification to different CNN representations. This is one of the first studies showing that an appropriate selection of the UQ method based on an understanding of the type of expected distribution shift, can further improve the performance of the task and the confidence estimation.

In addition, this detailed analysis of different types of natural distribution shift provides new insights into the robustness of the selected neural networks and their quantified uncertainty, the influence of different types of distribution shift, and performance differences on classification and regression tasks. ConvNeXt-Tiny is more robust throughout our experiments than ResNet-50 and EfficientNet-B0. Furthermore, runtime-efficient implementation of MC-Dropout, where dropout is applied to selected layers only, can increase the performance of the method. The selection of the proper runtime-efficient implementation depends on the type of expected distribution shift and on the task (classification or localization). For classification, After-BB-Dropout performs best, except for corner cases. LL-Dropout achieves higher accuracy for corner cases. For regression, LL-Dropout outperforms the other methods for confidence estimation and Head-Dropout performs best for localization accuracy. After-BB-Dropout is not recommended for localization.

Finally, our study shows that task performance and confidence estimation can significantly degrade under adverse weather, such as heavy rain or fog, and for corner cases and out-of-distribution data. It provides a quantification of these effects and a comparability of this performance reduction caused by different types of distribution shift. We hope this work inspires future work on robustness to natural distribution shift towards a more granular understanding of different types of data distortions, on the effects of those distortions on the different types of feature representations within a neural network, and on developing robust confidence estimation for safety-critical applications.